\documentclass[runningheads]{llncs}
\usepackage{graphicx}
\usepackage{amsmath,amssymb} 
\usepackage{color}

\usepackage{times}
\usepackage{epsfig}
\usepackage{graphicx}
\usepackage{amsmath}
\usepackage{amssymb}
\usepackage{bm}
\usepackage{subfigure}
\usepackage{booktabs} 
\usepackage{bbding}
\usepackage{amsmath,graphicx}
\usepackage{array}
\usepackage{caption}

\usepackage[width=122mm,left=12mm,paperwidth=146mm,height=193mm,top=12mm,paperheight=217mm]{geometry}
\begin{document}
\pagestyle{headings}
\mainmatter

\title{BSN: Boundary Sensitive Network for Temporal Action Proposal Generation} 

\titlerunning{Boundary Sensitive Network}

\authorrunning{Tianwei Lin \emph{et al.}}


\author{Tianwei Lin$^1$, Xu Zhao\thanks{ Corresponding author.}$^1$, Haisheng Su$^1$, Chongjing Wang$^2$, Ming Yang$^1$}


\institute{$^1$Department of Automation, Shanghai Jiao Tong University, China\\
	$^2$China Academy of Information and Communications Technology, China\\
	\email{ {wzmsltw,zhaoxu,suhaisheng,mingyang}@sjtu.edu.cn\\wangchongjing@caict.ac.cn}
}


\maketitle

\vspace{-0.5cm}

\begin{abstract}

Temporal action proposal generation is an important yet challenging problem, since temporal proposals with rich action content  are indispensable for analysing real-world videos with long duration and high proportion irrelevant content.
This problem requires methods not only generating proposals with precise temporal boundaries, but also retrieving proposals to cover truth action instances with high recall and high overlap using relatively fewer proposals.
To address these difficulties, we introduce an effective proposal generation method, named  Boundary-Sensitive Network (BSN), which adopts {\bf ``\emph{local to global}"} fashion.
{\bf \emph{Locally}}, BSN first locates temporal boundaries with high probabilities, then directly combines these boundaries as proposals.
{\bf \emph{Globally}},  with Boundary-Sensitive Proposal feature, BSN retrieves proposals by evaluating the confidence of whether a proposal contains an action within its region. 
We conduct experiments on two challenging datasets: ActivityNet-1.3 and THUMOS14, where BSN outperforms other state-of-the-art temporal action proposal generation methods with high recall and high temporal precision. 
Finally, further experiments demonstrate that by combining existing action classifiers, our method significantly improves the state-of-the-art temporal action detection performance.
\end{abstract}


\section{Introduction}


Nowadays, with fast development of  digital cameras and Internet, the number of videos is continuously booming,  making automatic  video content analysis methods widely required.
One major branch of video analysis is action recognition, which aims to classify manually trimmed video clips containing only one action instance.
However, videos in real scenarios are usually long, untrimmed and contain multiple action instances along with irrelevant contents. This problem requires algorithms for another challenging task: temporal action detection, which aims to detect action instances in untrimmed video including both temporal boundaries and action classes. It can be applied in many areas such as video recommendation and smart surveillance.

\begin{figure}[t]
\setlength{\abovecaptionskip}{-0.3cm} 
\setlength{\belowcaptionskip}{-0.5cm} 
\begin{center}
\begin{minipage}[b]{1.0\linewidth}
  \centering
  \centerline{\includegraphics[width=9.5cm]{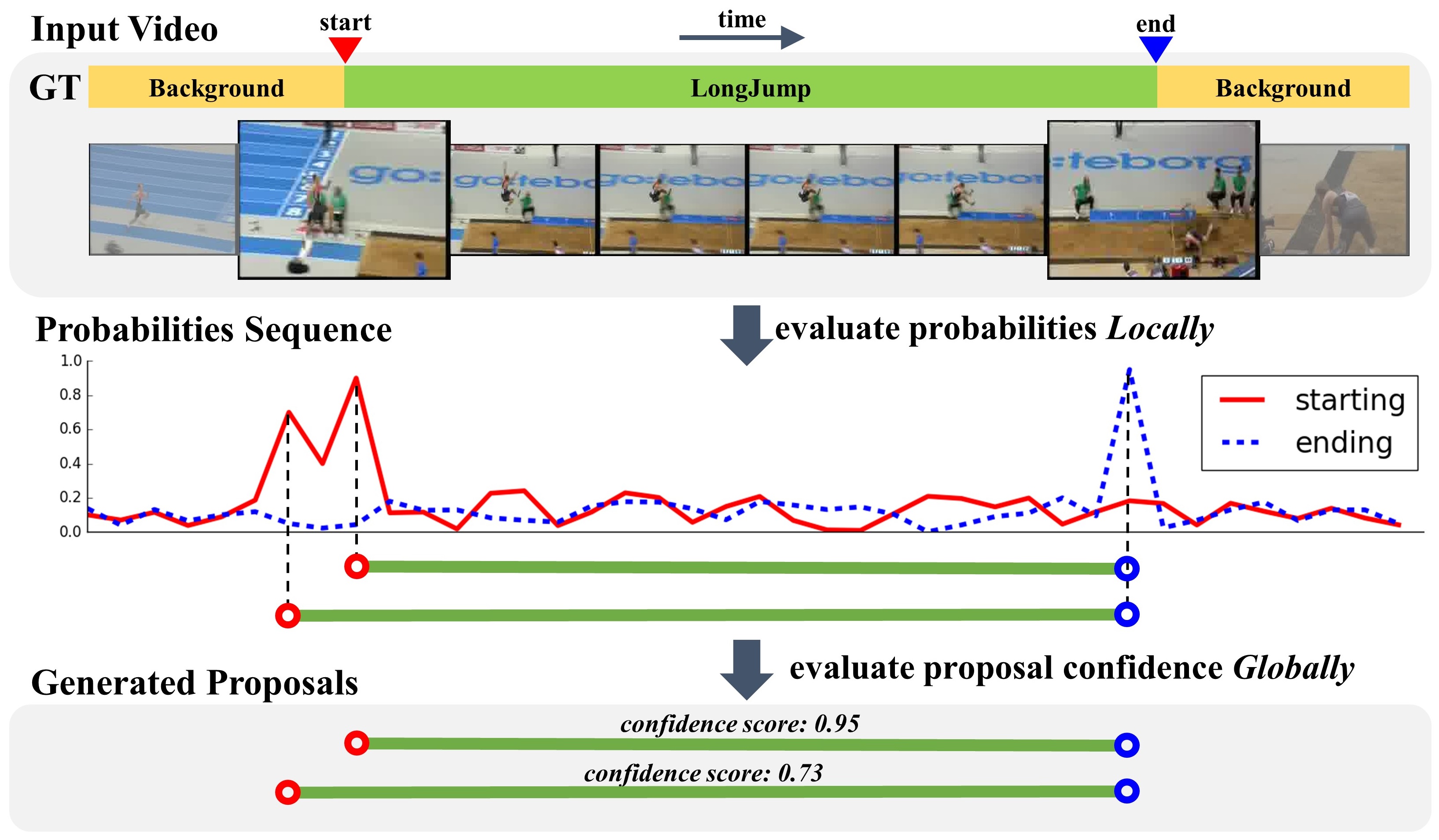}}
  \medskip
\end{minipage}
\end{center}
   \caption{Overview of our approach. Given an untrimmed video, (1) we evaluate boundaries and actionness probabilities of each temporal location and generate proposals based on boundary probabilities, and (2) we evaluate the confidence scores of proposals with proposal-level feature to get retrieved proposals.}
\label{fig_overview}
\end{figure}


Similar with object detection in spatial domain, temporal action detection task can be divided into two stages: proposal and classification. Proposal generation stage aims to generate temporal video regions which may contain action instances, and classification stage aims to classify classes of candidate proposals. 
Although  classification methods  have reached convincing performance,  the detection precision is still low in many benchmarks \cite{caba2015activitynet,jiang2014thumos}. 
Thus recently temporal action proposal generation has received much attention  \cite{sst_buch_cvpr17,fast_temporal_activity_cvpr16,escorcia2016daps,gao2017turn}, aiming to improve the detection performance by improving the quality of proposals.
High quality proposals should come up with two key properties: (1) proposals can cover truth action regions with both high recall and high temporal overlap, (2) proposals are retrieved so that high recall and high overlap can be achieved using fewer proposals to reduce the computation cost of succeeding steps.


To achieve high proposal quality, a proposal generation method should generate proposals with flexible temporal durations and precise temporal boundaries, then retrieve  proposals with reliable confidence scores, which indicate the probability of  a proposal containing an action instance.
Most recently proposal generation methods \cite{sst_buch_cvpr17,fast_temporal_activity_cvpr16,escorcia2016daps,shou2016action} generate proposals via sliding temporal windows of multiple durations in video with regular interval, then train a model to evaluate the confidence scores of generated proposals for proposals retrieving, while there is also method \cite{gao2017turn} making external boundaries regression.
However, proposals generated with pre-defined durations and intervals may have some major drawbacks: (1) usually not temporally precise; (2) not flexible enough to cover variable temporal durations of ground truth action instances, especially when the range of temporal durations is large.


To address these issues and generate high quality proposals, we propose the  Boundary-Sensitive Network (BSN), which adopts ``\emph{local to global}" fashion to locally combine high probability boundaries as proposals and globally retrieve candidate proposals using proposal-level feature as shown in Fig \ref{fig_overview}.
In detail, BSN generates proposals in three steps. 
{\bf First}, BSN evaluates the probabilities of each temporal location in video whether it is inside or outside, at or not at the boundaries of ground truth action instances, to generate starting, ending and actionness probabilities sequences as local information.
{\bf Second}, BSN generates proposals via directly  combining temporal locations with high starting and ending probabilities separately. Using this bottom-up fashion, BSN can generate proposals with flexible durations and precise boundaries. 
{\bf Finally}, using features composed by actionness scores within and around proposal, BSN retrieves proposals by evaluating the confidence of whether a proposal contains an action.  These proposal-level features offer global information for better evaluation.

In summary, the main contributions of our work are three-folds:


(1) We introduce a new architecture (BSN) based on ``\emph{local to global}" fashion to generate high quality temporal action proposals, which \emph{locally} locates high boundary probability locations to achieve precise proposal boundaries and \emph{globally} evaluates proposal-level feature to achieve reliable proposal confidence scores for retrieving.

(2) Extensive experiments demonstrate that our method achieves significantly better proposal quality than other state-of-the-art proposal generation methods, and can generate proposals in unseen action classes with comparative quality.

(3) Integrating our method with existing action classifier into detection framework leads to significantly improved performance on temporal action detection task.

\vspace{-0.1cm}

\section{Related work}

\noindent
{\bf Action recognition.}
Action recognition is an important branch of video related research areas and has been extensively studied. Earlier methods such as improved Dense Trajectory (iDT) \cite{dtf,wang2013action} mainly adopt hand-crafted features such as HOF, HOG and MBH. In recent years, convolutional networks are widely adopted in many works \cite{feichtenhofer2016convolutional,simonyan2014two,tran2015learning,wang2015towards} and have achieved great performance. Typically, two-stream network \cite{feichtenhofer2016convolutional,simonyan2014two,wang2015towards} learns appearance and motion features based on RGB frame and optical flow field separately. C3D network \cite{tran2015learning} adopts 3D convolutional layers to directly capture both appearance and motion features from raw frames volume.
Action recognition models can be used for extracting frame or snippet level visual features in long and untrimmed videos.

\noindent
{\bf Object detection and proposals.}
Recent years, the performance of object detection has been significantly improved with deep learning methods. R-CNN \cite{girshick2014rich} and its variations \cite{girshick2015fastt,ren2015faster} construct an important branch of object detection methods, which adopt ``detection by classifying proposals" framework.
For proposal generation stage, besides sliding windows \cite{felzenszwalb2010object}, earlier works also attempt to generate proposals by exploiting low-level cues such as HOG and Canny edge \cite{uijlings2013selective,zitnick2014edge}. Recently some methods \cite{ren2015faster,kuo2015deepbox,lin2016feature} adopt deep learning model to generate proposals with faster speed  and stronger modelling capacity.
In this work, we combine the properties of these methods via evaluating boundaries and actionness probabilities of each location using neural network and adopting ``\emph{local to global}" fashion to generate proposals with high recall and accuracy.

Boundary probabilities are also adopted in LocNet \cite{gidaris2016locnet} for revising the horizontal and vertical boundaries of existing proposals. Our method differs in (1) BSN aims to generate while LocNet aims to revise proposals and (2) boundary probabilities are calculated repeatedly for all boxes  in LocNet but only once for a video  in BSN.


\noindent
{\bf Temporal action detection and proposals.} 
Temporal action detection task aims to detect action instances in untrimmed videos including temporal boundaries and action classes, and can be divided into proposal and classification stages.
Most detection methods \cite{shou2016action,singh2016untrimmed,zhao2017temporal} take these two stages separately, while there is also method \cite{ssad,sstad} taking these two stages jointly.
For proposal generation, earlier works \cite{karaman2014fast,oneata2014lear,wang2014action} directly use sliding windows as proposals. Recently some methods \cite{sst_buch_cvpr17,fast_temporal_activity_cvpr16,escorcia2016daps,gao2017turn,shou2016action} generate proposals with  pre-defined temporal durations and intervals, and use multiple methods to evaluate the confidence score of proposals, such as dictionary learning \cite{fast_temporal_activity_cvpr16} and recurrent neural network \cite{escorcia2016daps}.
TAG method \cite{zhao2017temporal} adopts watershed algorithm to generate proposals with flexible boundaries and durations in \emph{local} fashion, but without \emph{global} proposal-level confidence evaluation for retrieving. 
In our work, BSN can generate proposals with flexible boundaries meanwhile reliable confidence scores for retrieving.

Recently temporal action detection method \cite{yuan2017temporal} detects action instances based on class-wise start, middle and end probabilities of each location. Our method is superior than \cite{yuan2017temporal} in two aspects: (1) BSN evaluates probabilities score using temporal convolution to better capture temporal information  and (2) ``\emph{local to global}" fashion adopted in BSN brings more precise boundaries and better retrieving quality.

\section{Our Approach}

\subsection{Problem Definition}

An untrimmed video sequence can be denoted as $X=\left \{ x_n \right \}_{n=1}^{l_v}$ with $l_v$  frames, where $x_n$ is the $n$-th frame in $X$.
Annotation of video $X$ is composed by a set of action instances $\Psi_g  = \left \{ \varphi  _n=\left (t_{s,n},t_{e,n}   \right ) \right \}_{n=1}^{N_g}$, where $N_g$ is the number of truth action instances in video  $X$, and $t_{s,n}$, $t_{e,n} $ are starting and ending time of action instance $\varphi_n$ separately.
Unlike  detection task, classes of action instances are not considered in temporal action proposal generation.
Annotation set $\Psi_g$ is used during training. During prediction, generated proposals set $\Psi_p$ should cover $\Psi_g$ with high recall and high temporal overlap.

\begin{figure*}
\setlength{\abovecaptionskip}{-0.4cm} 
\setlength{\belowcaptionskip}{-0.5cm} 
\begin{center}
\begin{minipage}[b]{1.0\linewidth}
  \centering
  \centerline{\includegraphics[width=12.2cm]{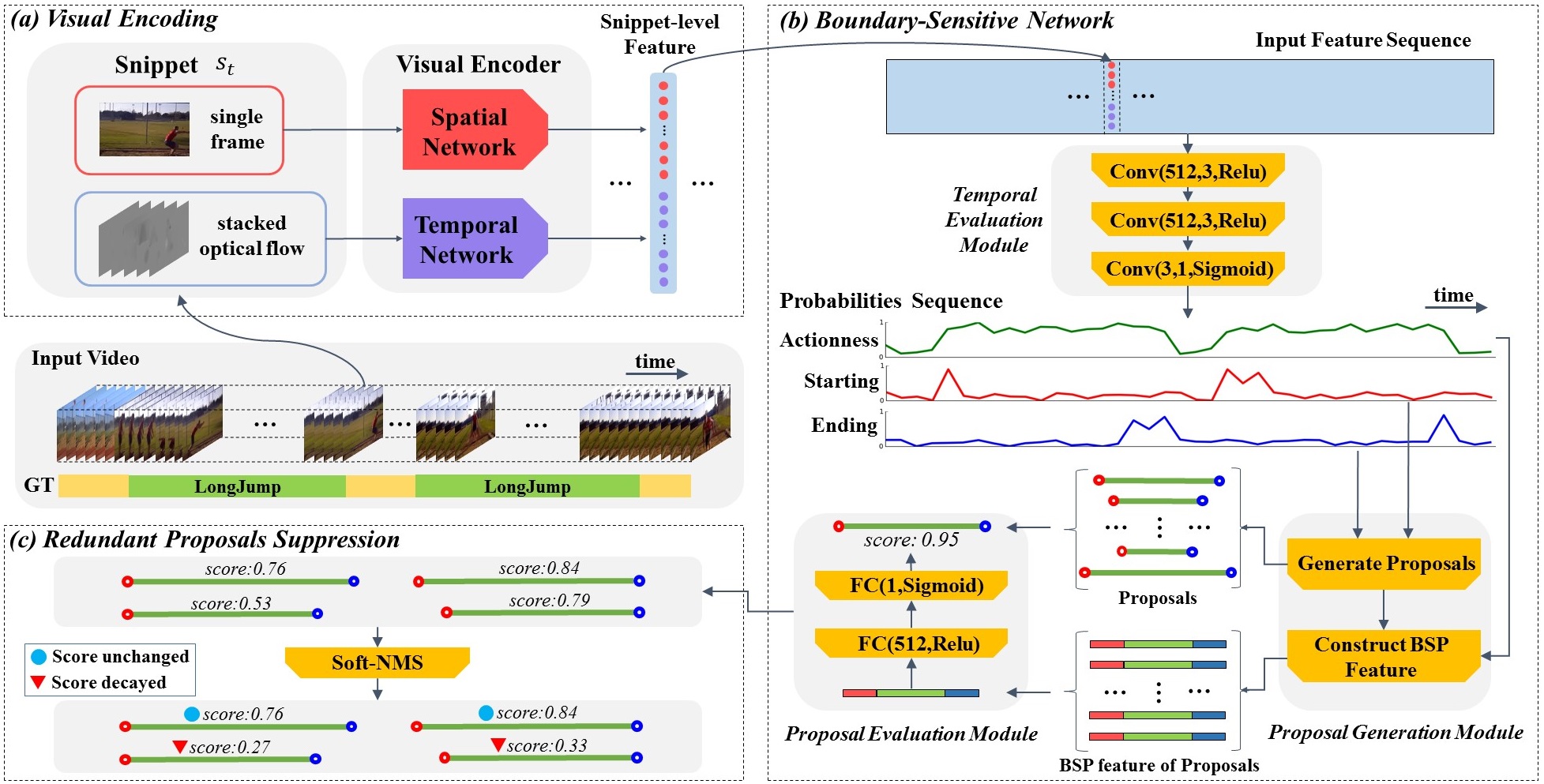}}
  \medskip
\end{minipage}
\end{center}
   \caption{The framework of our approach. (a) Two-stream network is used for encoding visual features in snippet-level. (b) The architecture of Boundary-Sensitive Network: \emph{temporal evaluation module} handles the input feature sequence, and evaluates starting, ending  and actionness probabilities of each temporal location; \emph{proposal generation module} generates proposals with high starting and ending probabilities, and construct Boundary-Sensitive Proposal (BSP) feature for each proposal; \emph{proposal evaluation module} evaluates confidence score of each proposal using BSP feature. (c) Finally, we use Soft-NMS algorithm  to suppress redundant proposals by decaying their scores. }
\label{fig_framework}
\end{figure*}

\subsection{Video Features Encoding}

To generate proposals of input video, first we need to extract feature to encode visual content of video. 
In our framework, we adopt two-stream network \cite{simonyan2014two} as visual encoder, since this architecture has shown great performance in action recognition task \cite{wang2016temporal} and has been widely adopted in temporal action detection and proposal generation tasks \cite{zhao2017temporal,ssad,gao2017cascaded}.
Two-stream network contains two branches: spatial network  operates on single RGB frame to capture appearance feature, and temporal network operates on stacked optical flow field to capture motion information.

To extract two-stream features, as shown in Fig \ref{fig_framework}(a), first we compose a snippets sequence $S=\left \{ s_n \right \}_{n=1}^{l_s}$ from video $X$, where $l_s$ is the length of snippets sequence.  A snippet $s_n=(x_{t_n}, o_{t_n})$ includes two parts: $x_{t_n}$ is the $t_n$-th RGB frame in $X$ and $o_{t_n}$ is stacked optical flow field derived around center frame $x_{t_n}$. 
To reduce the computation cost, we extract snippets with a regular frame interval $\sigma $, therefore  $l_s=l_v/\sigma$.
Given a snippet $s_n$, we concatenate output scores in top layer of both spatial and temporal networks to form the encoded feature vector $f_{t_n}=(f_{S,t_n},f_{T,t_n})$, where $f_{S,t_n}$, $f_{T,t_n}$ are output scores from spatial and temporal networks separately.  
Thus given a snippets sequence $S$ with length $l_s$, we can extract a feature sequence $F=\left \{ f_{t_n} \right \}_{n=1}^{l_s}$. These two-stream feature sequences are used as the input of BSN.

\subsection{Boundary-Sensitive Network}

To achieve high proposal quality with both precise temporal boundaries and reliable confidence scores, we adopt ``\emph{local to global}" fashion to generate proposals. In BSN, we first generate candidate boundary locations, then combine these locations as proposals and evaluate confidence score of each proposal with proposal-level feature.

\noindent
{\bf Network architecture.}
The architecture of BSN is presented in Fig \ref{fig_framework}(b), which contains three modules: temporal evaluation, proposal generation  and proposal evaluation. 
\emph{Temporal evaluation module} is a three layers temporal convolutional neural network, which  takes the two-stream feature sequences as input, and evaluates probabilities of each temporal location in video whether it is inside or outside, at or not at boundaries of ground truth action instances, to generate sequences of starting, ending and actionness probabilities respectively. 
\emph{Proposal generation module} first combines the temporal locations with separately high starting and ending probabilities as candidate proposals, then constructs Boundary-Sensitive Proposal (BSP) feature for each candidate proposal based on  actionness probabilities sequence. 
Finally, \emph{proposal evaluation module}, a multilayer perceptron model with one hidden layer, evaluates the confidence score of each candidate proposal based on BSP feature.
Confidence score and boundary probabilities of each proposal are fused as the final confidence score for retrieving. 

\noindent
{\bf Temporal evaluation module.}
The goal of temporal evaluation module is to evaluate starting, ending and actionness probabilities of each temporal location, where three binary classifiers are needed. In this module, we adopt temporal convolutional layers upon feature sequence, with good modelling capacity to capture local semantic information such as boundaries and actionness probabilities.

A temporal convolutional layer can be simply  denoted as $Conv(c_f,c_k,Act)$, where $c_f$, $c_k$ and $Act$ are filter numbers, kernel size and activation function of temporal convolutional layer separately. As shown in Fig \ref{fig_framework}(b), the temporal evaluation module can be defined as $Conv(512,3,Relu)\rightarrow Conv(512,3,Relu)\rightarrow Conv(3,1,Sigmoid)$, where the three layers have same stride size $1$. 
Three filters with sigmoid activation in the last layer are used as classifiers to generate starting, ending and actionness probabilities separately.
For convenience of computation, we divide feature sequence  into non-overlapped windows as the input of temporal evaluation module. 
Given a feature sequence $F$, temporal evaluation module can  generate three probability sequences $P_S=\left \{ p^s_{t_n} \right \}_{n=1}^{l_s}$, $P_E=\left \{ p^e_{t_n} \right \}_{n=1}^{l_s}$ and $P_A=\left \{ p^a_{t_n} \right \}_{n=1}^{l_s}$, where $p^s_{t_n}$, $p^e_{t_n}$ and $p^a_{t_n}$ are respectively starting, ending and actionness probabilities in time $t_n$.

\begin{figure}[t] 
\setlength{\abovecaptionskip}{-0.0cm} 
\setlength{\belowcaptionskip}{-0.5cm} 
\centering  
\makeatletter\def\@captype{figure}\makeatother 
\subfigure[Generate proposals]{
    \label{fig_with_crossover}
    \includegraphics[width=6.8 cm]{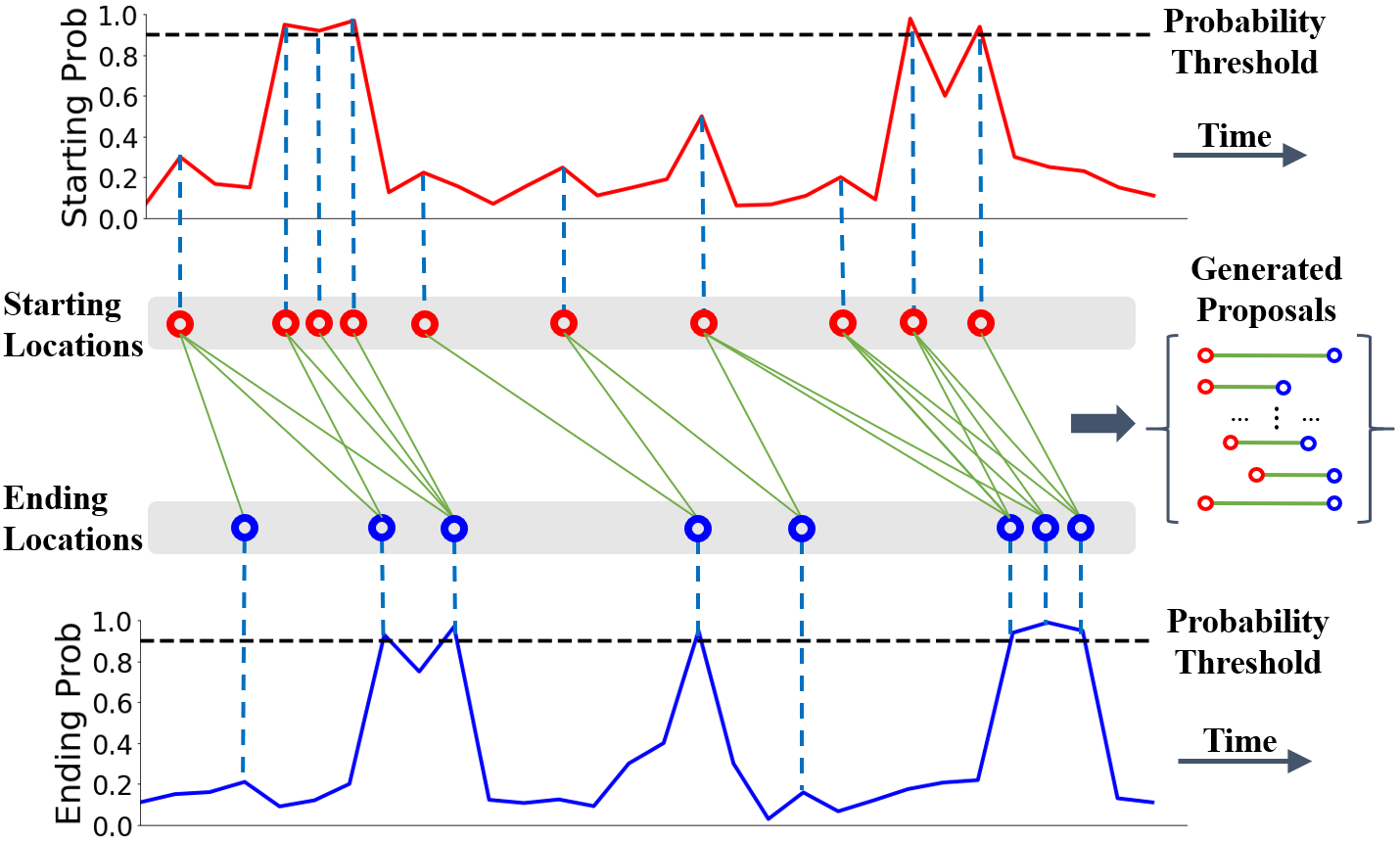}} 
\hspace{-0.2 in}  
\subfigure[Construct BSP feature]{
    \label{fig_without_crossover}
    \includegraphics[width=5.6 cm]{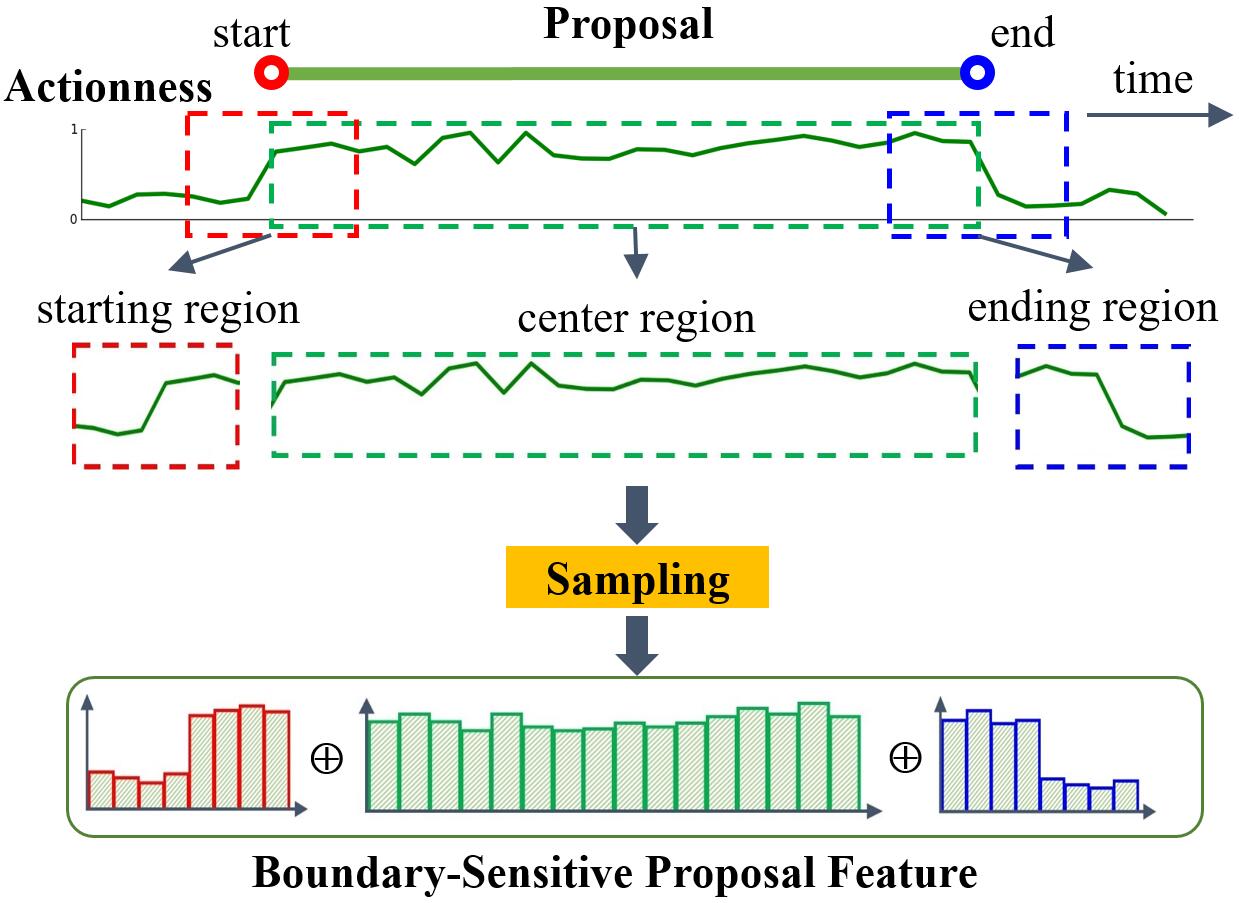}}
\caption{Details of proposal generation module. (a) Generate proposals. First, to generate candidate boundary locations, we choose temporal locations with high boundary probability or being a probability peak. Then, we combine candidate starting and ending locations as proposals when their duration satisfying  condition. 
(b) Construct BSP feature. Given a proposal and actionness probabilities sequence, we can sample actionness sequence in starting, center and ending regions of proposal to construct BSP feature.} 
\label{fig_pgm} 
\end{figure}

\noindent
{\bf Proposal generation module.}
The goal of proposal generation module is to generate candidate proposals and construct corresponding proposal-level feature.
We achieve this goal in two steps. First we locate temporal locations with high boundary probabilities, and combine these locations to form proposals. Then for each proposal, we construct Boundary-Sensitive Proposal (BSP) feature.

As shown in Fig \ref{fig_pgm}(a), to locate where an action likely to start, for starting probabilities sequence $P_S$, we record all temporal location $t_n$ where  $p^s_{t_n}$ (1) has high score: $p^s_{t_n}> 0.9$ or  (2) is a probability peak: $p^s_{t_n}> p^s_{t_{n-1}}$ and $p^s_{t_n}> p^s_{t_{n+1}}$.
These locations are  grouped into candidate starting locations set $B_S=\left \{ t_{s,i} \right \}_{i=1}^{N_S}$, where $N_S$ is the number of candidate starting locations. 
Using same rules, we can generate candidate ending locations set $B_E$ from ending probabilities sequence $P_E$.
Then, we generate temporal regions via combing each starting location $t_{s}$ from $B_S$ and each ending location $t_{e}$ from $B_S$. 
Any temporal region $\left [ t_s, t_e \right ]$ satisfying $ d=t_e- t_s \in [d_{min}, d_{max}]$ is denoted as a candidate proposal $\varphi$, where $d _{min}$ and $d_{max}$ are minimum and maximum durations of ground truth action instances in dataset. Thus we can get candidate proposals set $\Psi_p=\left \{ \varphi_i \right \}_{i=1}^{N_p}$, where $N_p$ is the number of proposals. 

To construct proposal-level feature as shown in Fig \ref{fig_pgm}(b), for a candidate proposal $\varphi$, we denote its  center region as $r_C=[t_s,t_e]$ and its starting and ending region as $r_S=[ t_s-d/5,t_s+d/5 ]$ and $r_E= [ t_e-d/5,t_e+d/5 ]$ separately. 
Then, we sample the actionness sequence $P_A$ within $r_c$ as $f_{c}^A$  by linear interpolation with 16 points. In starting and ending regions, we also sample actionness  sequence with 8 linear  interpolation points and get $f_{s}^A$ and $f_{e}^A$ separately. Concatenating these vectors, we can get Boundary-Sensitive Proposal (BSP) feature $f_{BSP}=(f_s^A$,$f_c^A$,$f_e^A)$ of proposal $\varphi$. 
BSP feature is highly compact and contains rich semantic information about corresponding proposal. 
Then we can represent a proposal as $\varphi=(t_s,t_e,f_{BSP})$.

\noindent
{\bf Proposal evaluation module.}
The goal of proposal evaluation module is to evaluate the confidence score of each proposal whether it contains an action instance within its duration using BSP feature. We adopt a simple multilayer perceptron model with one hidden layer as shown in Fig \ref{fig_framework}(b). Hidden layer with 512 units  handles the input of BSP feature $f_{BSP}$ with Relu  activation. The output layer outputs confidence score $p_{conf}$ with sigmoid activation, which estimates the overlap extent between candidate proposal and ground truth action instances. Thus, a generated proposal can be denoted  as $\varphi=(t_s,t_e,p_{conf},p^s_{t_s},p^e_{t_e})$, where $p^s_{t_s}$ and $p^e_{t_e}$ are starting and ending probabilities in $t_s$ and $t_e$ separately. These scores are fused to generate final score during prediction.

\subsection{Training of BSN }

In BSN, temporal evaluation module is trained to learn local boundary and actionness probabilities  from video features simultaneously. Then based on probabilities sequence generated by trained temporal evaluation module, we can generate proposals and corresponding BSP features and train the proposal evaluation module to learn the confidence score of proposals. The training details are introduced in this section.

\noindent
{\bf Temporal evaluation module.}
Given a video $X$, we compose a snippets sequence $S$ with length $l_s$ and extract feature sequence $F$ from it. 
Then we slide windows with length $l_w=100$ in feature sequence without overlap.
A window is denoted as $\omega =\left \{ F_{\omega}, \Psi_{\omega} \right \}$, where $F_{\omega}$ and $\Psi_{\omega}$ are feature sequence and annotations within the window separately. 
 For ground truth action instance $\varphi_g=( t_{s},t_{e} )$ in  $\Psi_{\omega}$, we denote its region as action region $r_{g}^a$ and its starting and ending region as $r_g^s=[ t_s-d_g/10,t_s+d_g/10 ]$ and $r_g^e= [ t_e-d_g/10,t_e+d_g/10 ]$ separately, where $d_g=t_e-t_s$.

Taking $F_{\omega}$ as input, temporal evaluation module generates probabilities sequence $P_{S,\omega}$, $P_{E,\omega}$ and $P_{A,\omega}$ with same length $l_w$. 
For each temporal location $t_n$ within $F_{\omega}$, we denote its region as $r _{t_n}=[ t_n-d_s/2,t_n+d_s/2 ]$ and get corresponding probability scores $p^s_{t_n}$, $p^e_{t_n}$ and $p^a_{t_n}$ from $P_{S,\omega}$, $P_{E,\omega}$ and $P_{A,\omega}$ separately, where $d_s=t_{n}-t_{n-1}$ is temporal interval between two snippets. 
Then for each  $r_{t_n}$, we calculate its $IoP$ ratio with $r_{g}^a$, $r_g^s$ and $r_g^e$ of all $\varphi_g$ in $\Psi_{\omega}$ separately, where $IoP$ is defined as the overlap ratio with groundtruth proportional to the duration of this proposal. Thus we can represent information of $t_n$ as $\phi_{n}=(p^a_{t_n},p^s_{t_n},p^e_{t_n},g^{a}_{t_n},g^{s}_{t_n},g^{e}_{t_n})$, where $g^{a}_{t_n}$, $g^{s}_{t_n}$, $g^{e}_{t_n}$ are maximum matching overlap $IoP$ of action, starting and ending regions separately.

Given a window of matching information as $\Phi_{\omega}=\left \{ \phi_n \right \}_{n=1}^{l_s}$, we can define training objective of this module as a three-task loss function. The overall loss function consists of actionness loss, starting loss and ending loss:

\begin{equation}
L_{TEM}=\lambda \cdot L_{bl}^{action}+L_{bl}^{start}+L_{bl}^{end} ,
\end{equation}
where $\lambda$ is the weight term and is set to 2 in BSN. We adopt the sum of binary logistic regression loss function $L_{bl}$ for all three tasks, which can be denoted as:

\begin{small}
\begin{equation}
L_{bl}=\frac{1}{l_w}\sum_{i=1}^{l_w} \left (\alpha^{+} \cdot b_i \cdot  log(p_i)+\alpha^{-} \cdot (1-b_i) \cdot log(1-p_i) \right ),
\end{equation}
\end{small}
where $b_i=sign(g_i-\theta_{IoP})$ is a two-values function for converting matching score $g_i$ to $\left \{0,1  \right \}$ based on threshold $\theta_{IoP}$, which is set to 0.5 in BSN.  Let $l^+=\sum g_i$ and $l^-=l_w-l^+$, we can set $\alpha^+=\frac{l_w}{l^-}$ and $\alpha^-=\frac{l_w}{l^+}$, which are used for balancing the effect of positive and negative samples during training.

\noindent
{\bf Proposal evaluation module.}
Using probabilities sequences generated by trained temporal evaluation module, we can generate proposals using proposal generation module: $\Psi_p  = \left \{ \varphi _n=(t_s,t_e,f_{BSP}) \right \}_{n=1}^{N_p}$. Taking $f_{BSP}$ as input, for a proposal $\varphi$,  confidence score $p_{conf}$ is generated by proposal evaluation module. Then we calculate its Intersection-over-Union (IoU)  with all $\varphi_g$ in $\Psi_g$, and denote the maximum overlap score as $g_{iou}$. Thus we can represent proposals set as $\Psi_p  = \left \{ \varphi _n =\left \{t_s,t_e,p_{conf},g_{iou}   \right \} \right \}_{n=1}^{N_p}$.
We split $\Psi_p$ into two parts based on $g_{iou}$: $\Psi_p^{pos}$ for $g_{iou}>0.7$ and  $\Psi_p^{neg}$ for $g_{iou}<0.3$. For data balancing, we take all proposals in $\Psi_p^{pos}$ and randomly sample the proposals in $\Psi_p^{neg}$ to insure the ratio between two sets be nearly 1:2.

The training objective of this module is a simple regression loss, which is used to train a precise confidence score prediction based on IoU overlap. We can define it as:
\begin{equation}
L_{PEM}=\frac{1}{N_{train}} \sum_{i=1}^{N_{train}}(p_{conf,i}-g_{iou,i})^2 ,
\end{equation}
where $N_{train}$ is the number of proposals used for training.

\subsection{Prediction and Post-processing}


During prediction, we use BSN with same procedures described in training to generate proposals set $\Psi_p  = \left \{  \varphi  _n=(t_s, t_e, p_{conf}, p^s_{t_s} ,  p^e_{t_e}  ) \right \}_{n=1}^{N_p}$, where   $N_p$ is the number of proposals.
To get final proposals set, we need to make score fusion to get final confidence score, then suppress redundant proposals based on these score.  

\noindent
{\bf Score fusion for retrieving.} 
To achieve better retrieving performance, for each candidate proposal $\varphi$,  we fuse its confidence score with its boundary probabilities by multiplication to get the final confidence score $p_{f}$:

\begin{equation}
p_{f}=p_{conf} \cdot p^s_{t_s} \cdot p^e_{t_e} .
\end{equation}

After score fusion, we can get generated proposals set $\Psi_p  = \left \{ \varphi  _n=(t_s,t_e,p_{f}  ) \right \}_{n=1}^{N_p}$, where $p_{f}$ is used for proposals retrieving. In section 4.2, we explore the recall performance with and without confidence score generated by proposal evaluation module.

\noindent
{\bf Redundant proposals suppression.} 
Around a ground truth action instance, we may generate multiple proposals with different temporal overlap. Thus we need to suppress redundant proposals to obtain higher recall with fewer proposals.

Soft-NMS \cite{softNMS} is a recently proposed non-maximum suppression (NMS) algorithm which suppresses redundant results using a score decaying function. First all proposals are sorted by their scores. Then proposal $\varphi_m$ with maximum score is used for calculating overlap IoU with other proposals, where scores of highly overlapped proposals is decayed. This step is recursively applied to the remaining proposals to  generate re-scored proposals set. 
The Gaussian decaying function of Soft-NMS can be denoted  as:

\vspace{-0.05cm}
\begin{equation}
p'_{f,i}=\left\{\begin{matrix}
p_{f,i}, & iou(\varphi_m,\varphi_i)<\theta\\ 
p_{f,i}\cdot e^{-\frac{iou(\varphi_m,\varphi_i)^2}{\varepsilon }}, & iou(\varphi_m,\varphi_i)\geq \theta
\end{matrix}\right. 
\end{equation}
where $\varepsilon $ is parameter of Gaussian function and $\theta$ is pre-fixed threshold. 
After suppression, we get the final proposals set $\Psi'_p  = \left \{ \varphi  _n=(t_s,t_e,p'_f  ) \right \}_{n=1}^{N_p}$.

\section{Experiments}

\subsection{Dataset and setup}
\noindent
{\bf Dataset.} {\bf ActivityNet-1.3} \cite{caba2015activitynet} is a large dataset for general temporal action proposal generation and detection, which contains 19994 videos   with 200 action classes annotated and was used in the ActivityNet Challenge 2016 and 2017. ActivityNet-1.3 is divided into training, validation and testing sets by ratio of 2:1:1. 
{\bf THUMOS14} \cite{jiang2014thumos} dataset contains 200 and 213 temporal annotated untrimmed videos with 20 action classes in validation and testing sets separately. The training set of THUMOS14 is the UCF-101 \cite{soomro2012ucf101}, which contains trimmed videos for action recognition task. 
In this section, we compare our method with state-of-the-art methods on both ActivityNet-1.3 and THUMOS14.

\noindent
{\bf Evaluation metrics.}
In temporal action proposal generation task, Average Recall (AR) calculated with multiple IoU thresholds is usually used as evaluation metrics. Following  conventions, we use IoU thresholds set $[0.5:0.05:0.95]$ in ActivityNet-1.3 and $[0.5:0.05:1.0]$ in THUMOS14. 
To evaluate the relation between recall and proposals number,  we evaluate AR with Average Number of proposals (AN) on both datasets, which is denoted as  AR@AN. 
On ActivityNet-1.3, area under the AR vs. AN curve (AUC) is also used as metrics, where AN varies from 0 to 100.

In temporal action detection task, mean Average Precision (mAP) is used as evaluation metric, where Average Precision (AP) is calculated on each action class respectively. On ActivityNet-1.3, mAP with IoU thresholds $\left \{0.5,0.75,0.95\right \}$ and average mAP with IoU thresholds set $[0.5:0.05:0.95]$ are used. On THUMOS14, mAP with IoU thresholds $\left \{0.3,0.4,0.5,0.6,0.7 \right \}$ is used.

\noindent
{\bf Implementation details.}
For visual feature encoding, we use the two-stream network \cite{simonyan2014two} with architecture described in \cite{xiong2016cuhk}, where BN-Inception network \cite{ioffe2015batch} is used as temporal network and ResNet network \cite{he2016deep} is used as spatial network. Two-stream network is implemented  using Caffe \cite{jia2014caffe} and pre-trained on ActivityNet-1.3 training set. During feature extraction,  the interval $\sigma $ of snippets is set to 16 on ActivityNet-1.3 and is set to 5 on THUMOS14.

On ActivityNet-1.3, since the duration of videos are limited, we follow \cite{lin2017temporal} to rescale the feature sequence of each video to new length $l_w =100$ by linear interpolation, and the duration of corresponding annotations to range [0,1].
In BSN, temporal evaluation module and proposal evaluation module are both implemented using Tensorflow \cite{abadi2016tensorflow}. On both datasets, temporal evaluation module is trained with batch size 16 and learning rate 0.001 for 10 epochs, then 0.0001 for another 10 epochs, and proposal evaluation module is trained with batch size 256 and same learning rate. 
For Soft-NMS, we set the threshold $\theta$ to 0.8 on ActivityNet-1.3 and 0.65 on THUMOS14 by empirical validation, while $\varepsilon $ in Gaussian function is set to 0.75 on both datasets.

\begin{figure*}[t]
\setlength{\abovecaptionskip}{-0.2cm} 
\setlength{\belowcaptionskip}{-0.7cm} 
\begin{center}
\begin{minipage}[b]{0.317\linewidth}
\centering
  {\includegraphics[width=\linewidth]{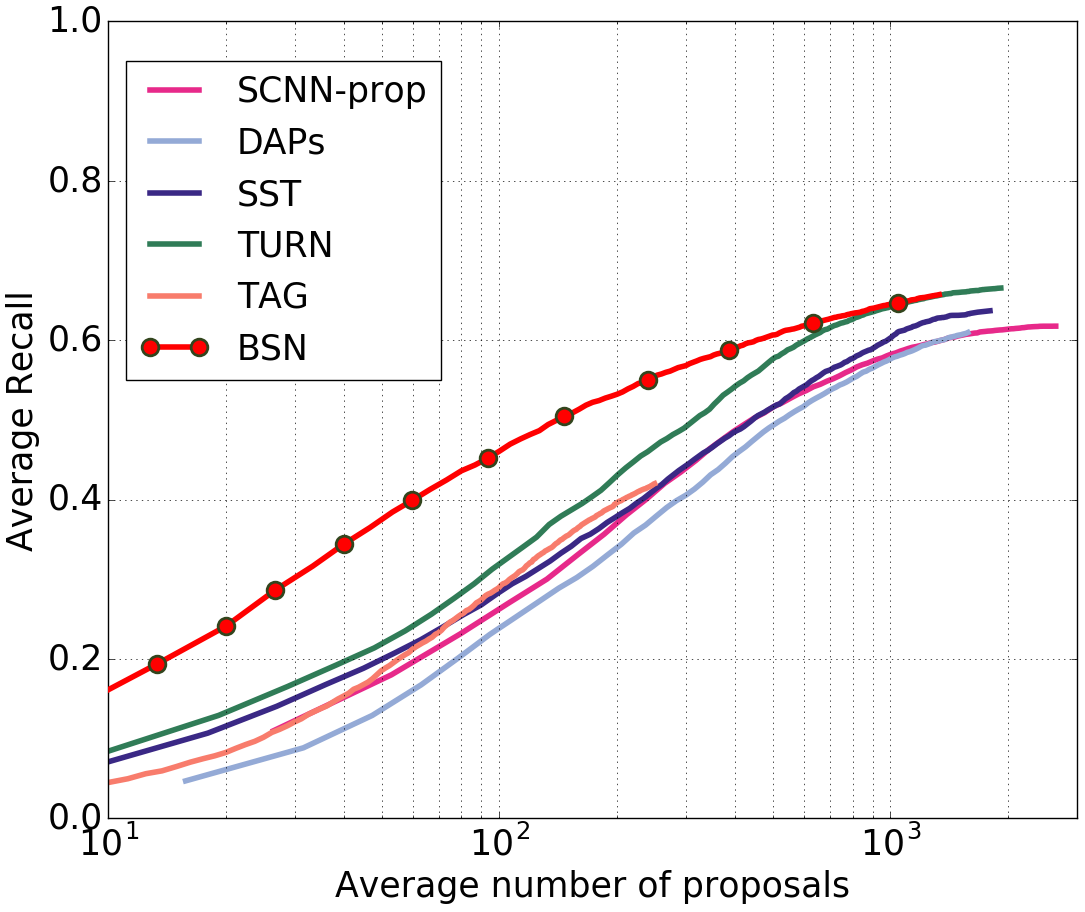}}
\end{minipage}
\hspace{-0.01 in}  
\begin{minipage}[b]{0.33\linewidth}
\centering
  {\includegraphics[width=\linewidth]{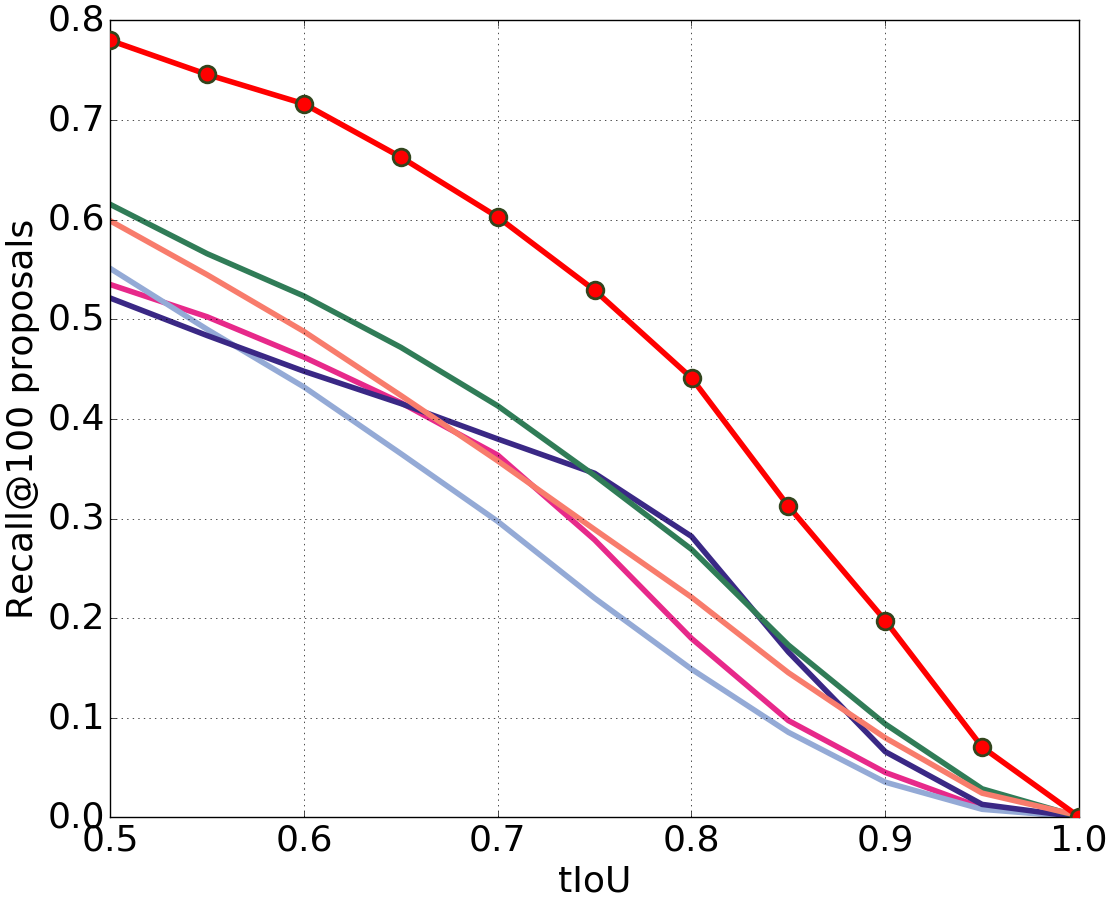}}
\end{minipage}
\hspace{-0.05 in}  
\begin{minipage}[b]{0.33\linewidth}
\centering
  {\includegraphics[width=\linewidth]{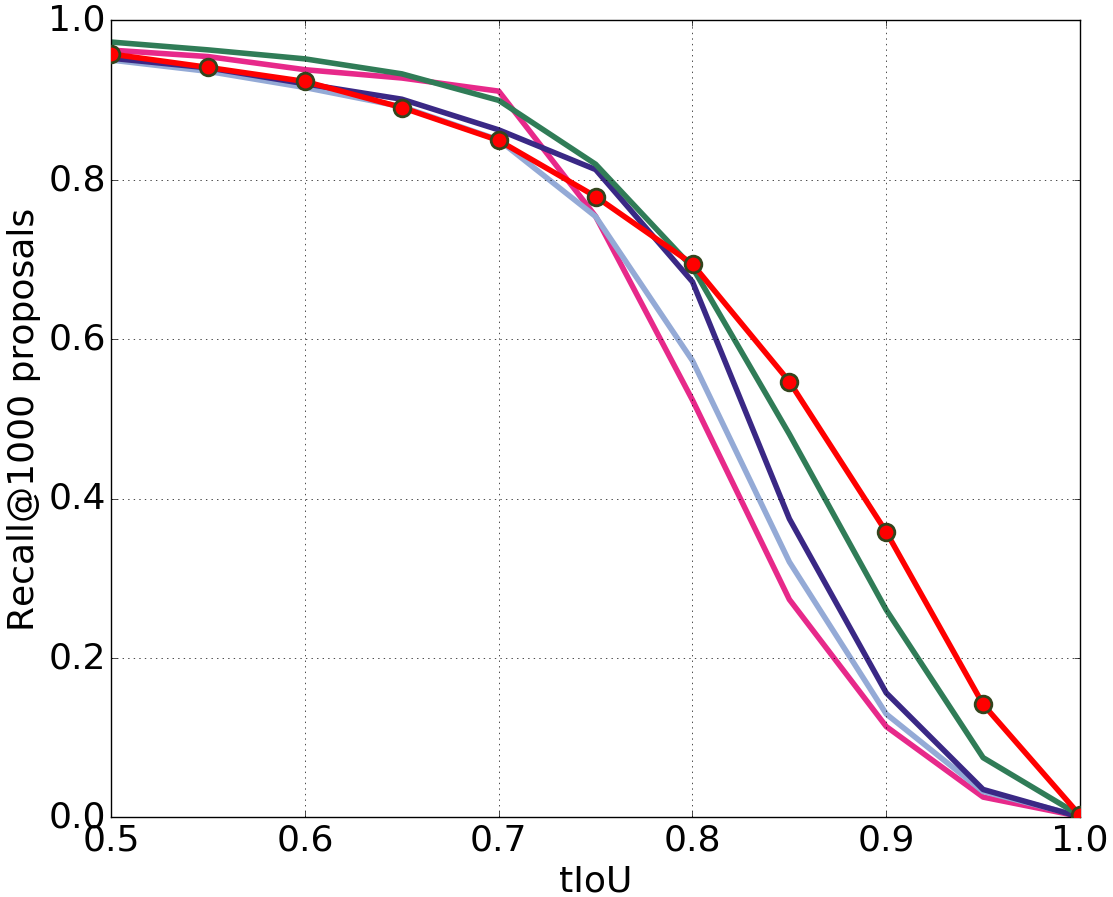}}
\end{minipage}
\end{center}
   \caption{Comparison of our proposal generation method with other state-of-the-art methods in THUMOS14 dataset. {\bf (left)} BSN can achieve significant performance gains with relatively few proposals. {\bf (center)} Recall with 100 proposals vs tIoU figure shows that with few proposals, BSN gets performance improvements in both low and high tIoU. {\bf (right)} Recall with 1000 proposals vs tIoU figure shows that with large number of proposals, BSN achieves improvements mainly while tIoU $> 0.8$. }
\label{fig_recall}
\end{figure*}

\vspace{-0.15cm}

\subsection{Temporal Proposal Generation}

Taking a video as input, proposal generation method aims to generate temporal proposals where action instances likely to occur. In this section, we compare our method with state-of-the-art methods and  make external experiments to verify effectiveness of BSN.

\noindent
{\bf Comparison with state-of-the-art methods.}
As aforementioned, a good proposal generation method should generate and retrieve proposals to cover ground truth action instances with \emph{high recall} and \emph{high temporal overlap}  using relatively few proposals. We evaluate these methods in two aspects.
 
First we evaluate the ability of our method to generate and retrieve proposals with high recall, which is measured by average recall with different number  of proposals (AR@AN) and area under AR-AN curve (AUC). We list the comparison results of ActivityNet-1.3 and THUMOS14 in Table \ref{table_comparison_1} and Table \ref{table_comparison_2} respectively, and plot the average recall against average number of proposals curve of THUMOS14 in Fig \ref{fig_recall} (left).
On THUMOS14, our method outperforms other state-of-the-art proposal methods when proposal number varies from 10 to 1000. Especially, when average number of proposals is 50, our method significantly improves average recall from $21.86\%$ to $37.46\%$ by $15.60\%$. On ActivityNet-1.3, our method outperforms other state-of-the-art proposal generation methods on both validation and testing set. 


Second, we evaluate the ability of our method to generate and retrieve proposals with high temporal overlap, which is measured by recall of multiple IoU thresholds. 
We plot the recall against IoU thresholds curve with 100 and 1000 proposals in Fig \ref{fig_recall} (center) and (right) separately.
Fig \ref{fig_recall} (center) suggests that our method achieves significant higher recall than other methods with 100 proposals when IoU threshold varied from 0.5 to 1.0. 
 And Fig \ref{fig_recall} (right) suggests that with 1000 proposals, our method obtains the largest recall improvements when IoU threshold is higher than 0.8. 

Furthermore,  we make some controlled experiments to confirm the contribution of BSN itself in Table \ref{table_comparison_2}.
For video feature encoding, except for two-stream network, C3D network \cite{tran2015learning} is also adopted in some works \cite{sst_buch_cvpr17,escorcia2016daps,gao2017turn,shou2016action}. For NMS method, most previous work adopt Greedy-NMS \cite{Dalal2005Histograms} for redundant proposals suppression.
%
%
Thus, for fair comparison, we train BSN with feature extracted by C3D network \cite{tran2015learning} pre-trained on UCF-101 dataset, then perform Greedy-NMS and Soft-NMS on C3D-BSN and original 2Stream-BSN respectively.
Results in Table \ref{table_comparison_2} show that (1) C3D-BSN still outperforms other C3D-based methods especially with small proposals number, (2)  Soft-NMS only brings small performance promotion  than Greedy-NMS, while Greedy-NMS also works well with BSN. 
These results suggest that the architecture of BSN itself  is the main reason for performance promotion rather than input feature and NMS method. 


\begin{table}[tbp]
\setlength{\abovecaptionskip}{-0.05cm} 
\setlength{\belowcaptionskip}{-0.1cm} 
\centering
\caption{ Comparison between our method with other state-of-the-art proposal generation methods on validation set of  ActivityNet-1.3 in terms of AR@AN and AUC. }
\small
\begin{tabular}{m{2.1cm}m{2.0cm}<{\centering}m{1.9cm}<{\centering}m{1.9cm}<{\centering}m{1.9cm}<{\centering}m{1.3cm}<{\centering}}
\toprule
Method  & Zhao et al. \cite{zhao2017temporal} & Dai et al. \cite{dai2017temporal}   & Yao et al. \cite{ghanem2017activitynet} & Lin et al. \cite{lin2017temporal}   &  BSN \\
\hline 
AR@100 (val)  	& 63.52  	& -		& - 		&  73.01		& {\bf 74.16}  \\
AUC (val) 		& 53.02  	& 59.58 	& 63.12 	&  64.40 	& {\bf 66.17	} \\
AUC (test) 		& -  		& 61.56	& 64.18 	&  64.80 	& {\bf 66.26}  \\
\bottomrule
\end{tabular}
\label{table_comparison_1}
\vspace{-0.5cm}
\end{table}

\begin{table}[tbp]
\setlength{\abovecaptionskip}{-0.05cm} 
\setlength{\belowcaptionskip}{0.1cm} 
\centering
\caption{ Comparison between our method with other state-of-the-art  proposal generation methods on  THUMOS14 in terms of AR@AN.} 
\begin{tabular}{m{1.9cm}<{\centering}m{3.1cm}m{1.2cm}<{\centering}m{1.2cm}<{\centering}m{1.2cm}<{\centering}m{1.2cm}<{\centering}m{1.2cm}<{\centering}}
\toprule
Feature & Method  		& @50 & @100  & @200 & @500 & @1000    \\
\hline 
C3D & DAPs \cite{escorcia2016daps} 		& 13.56	& 23.83 &  33.96 & 49.29 & 57.64   \\
C3D & SCNN-prop \cite{shou2016action} 	& 17.22 & 26.17 &  37.01 & 51.57 & 58.20   \\
C3D & SST \cite{sst_buch_cvpr17}			& 19.90 & 28.36  &  37.90 & 51.58 & 60.27   \\
C3D & TURN \cite{gao2017turn} 			& 19.63 & 27.96 &  38.34 & 53.52 & {\bf 60.75}  \\
\hline 
C3D & BSN + Greedy-NMS 	& 27.19 & 35.38 &  43.61 &  53.77  & 59.50   \\
C3D & BSN + Soft-NMS	& {\bf 29.58} & {\bf 37.38} &  {\bf 45.55} & {\bf 54.67} & 59.48   \\
\hline 
\hline 
2-Stream & TAG \cite{zhao2017temporal} 		& 18.55 & 29.00  &  39.61 & - & -  \\
Flow & TURN \cite{gao2017turn} 		& 21.86 & 31.89 & 43.02 & 57.63  & 64.17  \\
\hline 
2-Stream & BSN + Greedy-NMS & 	 35.41 & 43.55 &  52.23 & {\bf 61.35} & {\bf 65.10 } \\
2-Stream & BSN + Soft-NMS & {\bf 37.46} & {\bf 46.06} &  {\bf 53.21} & 60.64 &  64.52   \\
\bottomrule
\end{tabular}
\label{table_comparison_2}
\vspace{-0.4cm}
\end{table}




These results suggest the effectiveness of BSN. And BSN achieves the salient performance since it can generate proposals with (1) {\bf \emph{flexible temporal duration}} to cover ground truth action instances with various durations; (2) {\bf \emph{precise temporal boundary}} via learning starting and ending probability using temporal convolutional network, which brings high overlap between generated proposals and ground truth action instances; (3) {\bf \emph{reliable confidence score}} using BSP feature, which retrieves proposals properly so that high recall and high overlap can be achieved using relatively few proposals.
Qualitative examples on  THUMOS14 and ActivityNet-1.3 datasets are shown in Fig \ref{fig_vs}.

\begin{table}[tbp]
\centering
\caption{  Generalization evaluation of BSN on ActivityNet-1.3. \emph{Seen} subset: ``Sports, Exercise, and Recreation";  \emph{Unseen} subset: ``Socializing, Relaxing, and Leisure".  }
\begin{tabular}{m{5.5cm}m{1.5cm}<{\centering}m{1.5cm}<{\centering}m{1.5cm}<{\centering}m{1.5cm}<{\centering}}
\toprule
 & \multicolumn{2}{c}{\emph{Seen} (validation)} & \multicolumn{2}{c}{\emph{Unseen} (validation)}  \\
\hline   &  AR@100  & AUC & AR@100 & AUC \\
\hline 
BSN trained with \emph{Seen} + \emph{Unseen} (training)  	&  72.40  & 63.80 & 71.84 & 63.99 \\
BSN trained with \emph{Seen} (training)   			&  72.42  & 64.02 & {\bf 71.32}   & {\bf 63.38} \\
\bottomrule
\end{tabular}
\label{table_gene}
\vspace{-0.6cm}
\end{table}

\noindent
{\bf Generalizability of proposals.} Another key property of a proposal generation method  is the ability to generate proposals for unseen action classes. 
To evaluate this property, we choose two  semantically different action subsets on ActivityNet-1.3: ``Sports, Exercise, and Recreation" and ``Socializing, Relaxing, and Leisure" as \emph{seen} and \emph{unseen} subsets separately. \emph{Seen} subset contains 87 action classes with 4455  training and 2198 validation videos, and \emph{unseen} subset contains 38 action classes with 1903  training and 896 validation videos. 
To guarantee the experiment effectiveness, instead of two-stream network, here we adopt C3D network \cite{tran2017convnet} trained on Sports-1M dataset \cite{sports1m} for video features encoding. Using C3D feature, we train BSN with \emph{seen} and \emph{seen}+\emph{unseen} videos on training set separately, then evaluate both models on \emph{seen} and \emph{unseen} validation videos separately.
As shown in Table \ref{table_gene},  there is only slight performance drop in unseen classes, which demonstrates that BSN  has great generalizability and can learn a generic concept of temporal action proposal even in semantically different unseen actions.

\noindent
{\bf Effectiveness of modules in BSN.}
To evaluate the effectiveness of  temporal evaluation module (TEM) and proposal evaluation module (PEM) in BSN, we demonstrate experiment results of BSN with and without PEM in Table \ref{table_post}, where TEM is used in both results. These results show that: (1) using only TEM without PEM, BSN can also reach considerable recall performance over  state-of-the-art methods; (2) PEM can bring considerable further  performance promotion in BSN. These observations suggest that TEM and PEM are both effective and indispensable in BSN.

\noindent
{\bf Boundary-Sensitive Proposal feature.}
BSP feature is used in proposal evaluation module to evaluate the confidence scores of proposals. In Table \ref{table_post}, we also make ablation studies of the contribution of each component in BSP.  These results suggest that although BSP feature constructed from   boundary regions contributes less improvements than  center region, best recall performance is achieved while PEM is trained with BSP constructed from both boundary and center region.

\vspace{-0.25cm}

\subsection{Action Detection with  Our Proposals}

To further evaluate the quality of proposals generated by BSN, we put BSN proposals into ``detection by classifying proposals" temporal action detection framework with state-of-the-art action classifier, where  temporal boundaries of detection results are provided by our proposals.
On ActivityNet-1.3, we use top-2 video-level class generated by classification model \cite{zhao2017cuhk}\footnote{Previously, we adopted classification results from result files of \cite{wang2016uts}. Recently we found that the classification accuracy of these results are unexpected high. Thus we replace it with classification results of \cite{zhao2017cuhk} and updated all related experiments accordingly.}  for all proposals in a video. 
On THUMOS14, we use top-2 video-level classes generated by UntrimmedNet \cite{wang2017untrimmednets} for proposals generated by BSN and other methods.
Following  previous works,  on THUMOS14, we also implement SCNN-classifier on BSN proposals for proposal-level classification and  adopt Greedy NMS  as \cite{shou2016action}.
We use 100 and 200 proposals per video on ActivityNet-1.3  and THUMOS14 datasets separately.

\begin{table}[tbp]
\setlength{\abovecaptionskip}{0.1cm} 
\small
\centering
\caption{ Study of effectiveness of modules in BSN and  contribution of components in BSP feature on THUMOS14, where PEM is trained with BSP feature constructed by   \emph{Boundary} region ($f_s^A,f_e^A$) and   \emph{Center} region ($f_c^A$) independently and jointly.}
\begin{tabular}{p{2.7cm}p{1.3cm}<{\centering}p{1.3cm}<{\centering}p{1.2cm}<{\centering}p{1.2cm}<{\centering} p{1.2cm}<{\centering}p{1.2cm}<{\centering}p{1.2cm}<{\centering}}
\toprule
					& \emph{Boundary} 	& \emph{Center} 	& @50	& @100 	& @200	& @500	& @1000 \\
 \hline
 BSN without PEM 	&			&			& 30.72	& 40.52 	& 48.63	& 57.78	& 63.04\\
 \hline
 					&\Checkmark 	& 			& 35.61	& 44.86	& 52.46	& 60.00	& 64.17\\
 BSN with PEM 		&  			&\Checkmark	& 36.80	& 45.65	& 52.63	& 60.18	& 64.22\\
 					&\Checkmark	&\Checkmark	& {\bf 37.46}	& {\bf 46.06	}& {\bf 53.21}	& {\bf 60.64}	& {\bf 64.52}\\
\bottomrule
\end{tabular}
\label{table_post}
\vspace{-0.6cm}
\end{table}

\begin{table}[tbp]
\setlength{\abovecaptionskip}{0.1cm} 
\centering
\caption{Action detection results on validation and testing set of ActivityNet-1.3 in terms of mAP@$tIoU$ and average mAP, where our proposals are combined with video-level classification results generated by \cite{zhao2017cuhk}.  }
\small
\begin{tabular}{p{2.8cm}p{1.4cm}<{\centering}p{1.4cm}<{\centering}p{1.4cm}<{\centering}p{1.5cm}<{\centering}p{1.6cm}<{\centering}}
\toprule
 & \multicolumn{4}{c}{validation} & testing  \\
\hline
Method  & 0.5  &  0.75  & 0.95  & Average  & Average  \\
\hline 
Wang et al. \cite{wang2016uts}    & 42.28 & 3.76  & 0.05   & 14.85 & 14.62 \\
SCC \cite{heilbron2017scc}   & 40.00 & 17.90  & 4.70   & 21.70 & 19.30 \\
CDC \cite{shou2017cdc}    & 43.83  & 25.88  & 0.21   & 22.77  & 22.90 \\
TCN \cite{dai2017temporal} & - & - & - & - & 23.58\\
SSN \cite{xiong2017pursuit}    & 39.12 & 23.48  & 5.49  & 23.98 & 28.28 \\
Lin et al. \cite{lin2017temporal} & 44.39   & 29.65  & 7.09  & 29.17 & 32.26 \\
\hline
BSN + \cite{zhao2017cuhk} & {\bf 46.45 }   & {\bf 29.96}  & {\bf 8.02}  & {\bf 30.03 } & {\bf 32.84 } \\
\bottomrule
\end{tabular}
\label{table_detection_anet}
\normalsize
\vspace{-0.1cm}                                                              
\end{table}

\begin{table}[tbp]
\setlength{\abovecaptionskip}{0.1cm} 
\setlength{\belowcaptionskip}{-0.3cm} 
\centering
\caption{Action detection results on testing set of THUMOS14 in terms of mAP@$tIoU$, where classification results generated by UntrimmedNet \cite{wang2017untrimmednets} and  SCNN-classifier \cite{shou2016action} are combined with proposals generated by BSN and other  methods. }

\begin{tabular}{p{2.4cm}p{1.6cm}p{1.2cm}<{\centering}p{1.2cm}<{\centering}p{1.2cm}<{\centering}p{1.2cm}<{\centering}p{1.2cm}<{\centering}}
\toprule
\multicolumn{7}{c}{  Action Detection Methods }  \\
\hline
\multicolumn{2}{l}{  Detection Method } & 0.7 & 0.6 & 0.5 & 0.4 & 0.3  \\
\hline  
\multicolumn{2}{l}{  SCNN \cite{shou2016action} } & 5.3 & 10.3 &  19.0 & 28.7 & 36.3 \\
\multicolumn{2}{l}{  SMS \cite{yuan2017temporal} } & - & - & 17.8 &  27.8 &  36.5 \\
\multicolumn{2}{l}{  CDC \cite{shou2017cdc} } & 8.8 & 14.3 &  24.7 & 30.7 & 41.3  \\
\multicolumn{2}{l}{  SSAD \cite{ssad} } & 7.7 & 15.3 & 24.6 &  35.0 &  43.0 \\
\multicolumn{2}{l}{  TCN \cite{dai2017temporal} } & 9.0 & 15.9 & 25.6 &  33.3 &  - \\
\multicolumn{2}{l}{  R-C3D \cite{xu2017r} } & 9.3 & 19.1 &  28.9 & 35.6 & 44.8 \\
\multicolumn{2}{l}{  SS-TAD \cite{sstad} } & 9.6 & - &  29.2 & - & 45.7 \\
\multicolumn{2}{l}{  SSN  \cite{xiong2017pursuit} } & - & - & 29.1 & 40.8 & 50.6 \\
\multicolumn{2}{l}{  CBR \cite{gao2017cascaded} } & 9.9 & 19.1 &  31.0 & 41.3 & 50.1 \\
\hline 
\multicolumn{7}{c}{   Proposal Generation Methods + Action Classifier }  \\
\hline
Proposal method & Classifier & 0.7 & 0.6 & 0.5 & 0.4 & 0.3  \\
\hline
SST \cite{sst_buch_cvpr17} & SCNN-cls 	& - & - & 23.0 & - &  -\\
TURN \cite{gao2017turn} & SCNN-cls 		& 7.7 & 14.6 & 25.6 & 33.2 &  44.1\\
SST \cite{sst_buch_cvpr17} & UNet 		& 4.7 & 10.9 & 20.0 & 31.5 &  41.2\\
TURN \cite{gao2017turn} & UNet 			& 6.3 & 14.1 & 24.5 & 35.3 &  46.3\\
\hline
BSN & SCNN-cls & 15.0 & 22.4 & 29.4 & 36.6 &   43.1\\
BSN & UNet & {\bf 20.0} & {\bf 28.4} & {\bf 36.9} & {\bf 45.0} &  {\bf 53.5}\\
\bottomrule
\end{tabular}
\label{table_detection_thumos}
\normalsize
\end{table}

\begin{figure}[t]
\setlength{\abovecaptionskip}{-0.3cm} 
\setlength{\belowcaptionskip}{-0.3cm} 
\begin{center}
\begin{minipage}[b]{1.0\linewidth}
  \centering
  \centerline{\includegraphics[width=12.2cm]{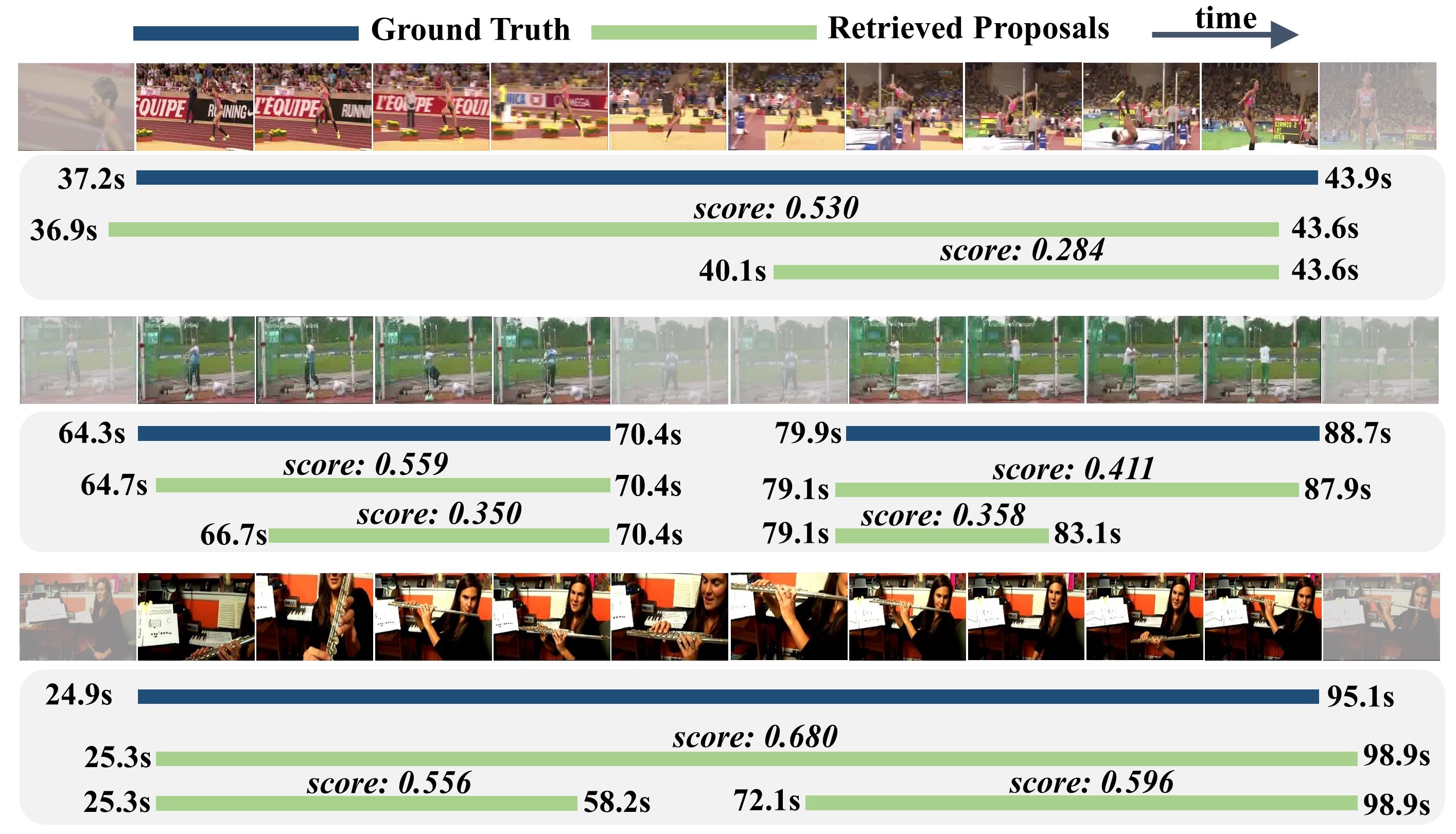}}
  \medskip
\end{minipage}
\end{center}
   \caption{Qualitative examples of proposals generated by BSN on THUMOS14 (top and middle) and ActivityNet-1.3 (bottom), where proposals are retrieved using  post-processed confidence score. }
\label{fig_vs}
\vspace{-0.2cm}
\end{figure}

The comparison results of ActivityNet-1.3 shown in Table \ref{table_detection_anet} suggest that  detection framework based on our proposals outperforms other state-of-the-art methods.
The comparison results of THUMOS14 shown in Table \ref{table_detection_thumos} suggest that 
(1) using same action classifier, our method achieves significantly better performance than other proposal generation methods;
(2) comparing  with  proposal-level classifier \cite{shou2016action}, video-level classifier \cite{wang2017untrimmednets} achieves better performance on BSN proposals and worse performance  on \cite{sst_buch_cvpr17} and \cite{gao2017turn} proposals, which indicates that confidence scores generated by BSN are more reliable than scores generated by proposal-level classifier, and are reliable enough for retrieving detection results in action detection task;
(3) detection framework based on our proposals significantly outperforms  state-of-the-art action detection methods, especially when the overlap threshold is high. 
These results confirm that proposals generated by BSN have high quality and work generally well in detection frameworks. 




\section{Conclusion}

In this paper, we have introduced the Boundary-Sensitive Network (BSN) for temporal action proposal generation. Our method can generate proposals with flexible durations and precise boundaries via directly combing locations with high boundary probabilities, and make accurate  retrieving via evaluating proposal confidence score with proposal-level features. Thus BSN can achieve high recall and high temporal overlap with relatively few proposals.
In experiments, we demonstrate that BSN significantly outperforms other state-of-the-art proposal generation methods on both THUMOS14 and ActivityNet-1.3 datasets. And BSN can significantly improve  the detection performance when used as the proposal stage of a full detection framework.


\newpage

\bibliographystyle{splncs}
\bibliography{egbib}
\end{document}